\documentclass[11pt, a4]{article}
\usepackage{coling2016}
\usepackage{times}
\usepackage{url}
\usepackage{latexsym}
\usepackage{subfigure}
\usepackage{graphicx}
\usepackage{multirow}
\usepackage{amssymb,amsfonts}
\makeatletter
\newcommand{\@BIBLABEL}{\@emptybiblabel}
\newcommand{\@emptybiblabel}[1]{}
\makeatother
\usepackage[hidelinks]{hyperref}

\long\def\eat#1{\ignorespaces}

\def\dnrm#1{\mbox{$_{\hbox{\scriptsize #1}}$}}

\title{Simple Question Answering by Attentive Convolutional Neural Network\thanks{\enspace This work was conducted during the first author's internship at IBM Watson Group.}}

\author{Wenpeng Yin$^*$, \enspace Mo Yu$^\dag$, \enspace Bing Xiang$^\dag$,\\
$^*$Center for Information and Language Processing\\
LMU Munich, Germany\\
{\tt wenpeng@cis.lmu.de}\\\And
Bowen Zhou$^\dag$, \enspace Hinrich Sch\"{u}tze$^*$\\
$^\dag$IBM Watson\\
Yorktown Heights, NY, USA\\
{\tt \{yum,bingxia,zhou\}@us.ibm.com}\\}

\date{}
\def\figref#1{Figure~\ref{fig:#1}}

\def\seclabel#1{\label{sec:#1}\label{p:#1}}
\def\eqref#1{Eq.~\ref{eqn:#1}}

\def\eqlabel#1{\label{eqn:#1}}

\newcounter{notecounter}
\newcommand{\enotesoff}{\long\gdef\enote##1##2{}}
\newcommand{\enoteson}{\long\gdef\enote##1##2{{
\stepcounter{notecounter}
{\large\bf
\hspace{1cm}\arabic{notecounter} $<<<$ ##1: ##2
$>>>$\hspace{1cm}}}}}
\enoteson
\enotesoff
\begin{document}
\maketitle
\begin{abstract}
This work focuses on answering single-relation factoid
questions over Freebase. Each question can acquire the
answer from a single fact of form (subject, predicate,
object) in Freebase.  This task, simple question answering
(SimpleQA), can be addressed via a two-step pipeline: entity
linking and fact selection. In fact selection, we match the
\emph{subject entity in a fact candidate} with the entity mention in
the question by a \emph{character-level} convolutional neural
network (char-CNN), and match the \emph{predicate in that fact} with
the question by a \emph{word-level} CNN (word-CNN). This work makes two
main contributions. (i) A simple and effective entity linker
over Freebase is proposed. Our entity linker outperforms the
state-of-the-art entity linker over SimpleQA task.  \footnote{\emph{We release our entity linking results at}: \url{https://github.com/Gorov/SimpleQuestions-EntityLinking}} (ii) A novel attentive maxpooling is stacked over word-CNN, so that the
predicate representation can be matched with the predicate-focused
question representation more effectively. Experiments
show that our system sets new state-of-the-art in this task.
\end{abstract}

\section{Introduction}\label{sec:intro}
\blfootnote{
    %
    % for review submission
    %
    % \hspace{-0.65cm}  % space normally used by the marker
    % Place licence statement here for the camera-ready version, see
    % Section~\ref{licence} of the instructions for preparing a
    % manuscript.
    %
    % % final paper: en-uk version 
    %
    % \hspace{-0.65cm}  % space normally used by the marker
    % This work is licenced under a Creative Commons 
    % Attribution 4.0 International Licence.
    % Licence details:
    % \url{http://creativecommons.org/licenses/by/4.0/}
    % 
    % final paper: en-us version 
    
    %\hspace{-0.65cm}  % space normally used by the marker
    This work is licensed under a Creative Commons 
    Attribution 4.0 International License.
    License details:
    \url{http://creativecommons.org/licenses/by/4.0/}
}

Factoid question answering (QA) over knowledge bases such as
Freebase \cite{bollacker2008freebase} has been intensively
studied recently
(e.g.,  \newcite{bordes2014question}, \newcite{yao2014freebase}, \newcite{bast2015more}, \newcite{yih2015semantic}, \newcite{xu2016enhancing}).
Answering a  question can require reference  to multiple
related facts 
in Freebase or reference to a single fact.
This work studies  simple question answering
(SimpleQA) based on the \emph{SimpleQuestions} benchmark
\cite{bordes2015large} in which answering a question does not
require reasoning over multiple facts. Single-relation factual questions
are the most common type of question observed in various
community QA sites \cite{fader2013paraphrase} and  in
search query logs. 
Even though this task is called ``simple'', it is in reality
not simple at all and far from solved.

In SimpleQA, a question, such as ``what's the hometown
of Obama?'', asks a single and direct topic of an
entity. In this example, the  entity is ``Obama'' and
the topic is hometown.  So our task is reduced to 
finding one fact (subject, predicate, object) in Freebase that answers
the question, which roughly means the \emph{subject} and \emph{predicate} are the best
matches for the topical entity ``Obama'' and for the topic
description ``what's the hometown of'', respectively.
Thus, we aim to design a method that 
picks a fact from Freebase, so that this fact
matches the question best. This procedure resembles   answer
selection \cite{yu2014deep} in which a system, given a question, is asked to
choose the best answer from a list of candidates. In this work, we formulate the SimpleQA task as a
fact selection problem and the key issue lies in the
system design for how to match a fact candidate to the question.

The first obstacle is that Freebase has an overwhelming
number of facts. A common and effective way is to first
conduct entity linking of a question over Freebase, so that
only a small subset of facts remain as candidates. Prior work
achieves entity linking by searching word $n$-grams of a
question among all entity names
\cite{bordes2015large,golub2016character}. Then, facts
whose subject entities match those $n$-grams are kept.  Our
first contribution in this work is to present a simple while
effective entity linker  to this task. Our
entity linker first uses each word of a question (or of an entity mention in the question) to search
in the entity vocabulary, all entities are kept if their names
\emph{contain one of the query words}. Then, we design three simple
factors to give a raw ranking score for each entity
candidate: (i) the ratio of words in the entity name that are
covered by the question; (ii) the ratio of words in the
question that are covered by the entity name; (iii) the position
of the entity mention in the question. We choose top-$N$
ranked entities as candidates. Our entity linker does not
consider the semantics or topic of an entity; it considers
only the string surface. Nevertheless, experiments  show
that these three factors are the basis for a top-performing entity
linker for SimpleQA.

Based on entity linking results, we consider each fact as
a fact candidate that has
one of the entity candidates as subject.
Then our system solves the task of
\emph{fact selection}, i.e., matching the
question with each fact candidate and picking the best one. Our system is built
based on two observations. (i) 
Surface-form match between a subject entity and its mention in the
question provides more straight-forward and effective clue than their semantic match. For example,
``Barack Obama'' matches with ``Obama'' in surface-form,
which acts as a fundamental indicator that the corresponding
fact and the question are possibly about the same
``Obama''. (ii) Predicate in a fact is a paraphrase of
the question's pattern where we define the \emph{pattern} to be the
topic asked by the question
about the entity, and
represent it
as the question in which the
entity mention 
has been 
replaced by a special
symbol. Often the predicate corresponds to a keyword or a rephrased
token of the pattern,  this means we need to create a
flexible  model to handle this
relationship.

These observations motivate us to include
two kinds of convolutional neural networks (CNN,
\newcite{lecun1998gradient})
in our deep learning system. (i) A character-level CNN
(\emph{char-CNN}) that models the match between an Freebase entity and
its mention in the question on surface-form. We consider CNN
over character-level rather than the commonly-used
word-level, so that the generated representation  is
more robust even in the presence of typos, spaces and
other character violations. (ii) A word-level CNN
(\emph{word-CNN}) with attentive maxpooling that learns the
match of the Freebase predicate  with the 
question's pattern. A Freebase predicate  is a predefined relation, 
mostly consisting of a few
words:  ``place of birth'', ``nationality'', ``author editor'' etc.
 In contrast, a pattern is highly variable in length and
 word choice, i.e., the subsequence of the question that 
represents the predicate in a question can take many
different forms.
Convolution-maxpooling slides a window over the input and
identifies the best matching subsequence for a task, using a
number of filters that support flexible matching. Thus, 
convolution-maxpooling is an appropriate method for finding
the pattern subsequence that best matches the predicate description. 
\emph{We add attention to this basic operation of convolution-maxpooling.}
Attentions are guided by the predicate over all $n$-gram phrases in the
pattern,  finally system pools phrase features by considering the feature values as well as the attentions towards those features.
\emph{Phrases more
similar to the predicate, i.e., with higher attention values,
will be selected with higher probability than other phrases to represent the pattern}.\footnote{Surface-form entity linking has
  limitations in candidate collection as some entities have
  the same names. We tried another word-CNN to match the
  pattern to the entity description provided by Freebase,
  but no improvement is observed.}

% \eat{
% Traditional maxpooling
% over convolution outputs extract features from each output
% hidden vector with the same probability. Intuitively,
% predicate representation should be more close to some parts
% of the question than other parts. Hence, we prefer that the
% maxpooling operation can extract features from the hidden
% representations of similar parts with higher
% probability. Motivated, we come up with a scheme to learn
% the attentions of predicate over all $n$-gram phrases in the
% question, and finally do maxpooling over those phrase
% representations with different probabilities. Phrases more
% similar to the predicate, i.e., with higher attention values,
% will contribute more in the final question's representation
% than other phrases.\footnote{Surface-form entity linking has
%   limitations in candidate collection, as some entities have
%   the same names. We tried another word-CNN to match the
%   question to the entity description provided by Freebase,
%   but no improvement is observed.}
% }

Our overall approach is for the entity linker to identify top-$N$
entity candidates
for a question. All facts that contain one of these
entities as subject are then the fact search space for this question.
Char-CNN and
word-CNN decompose each question-fact match into
an
entity-mention
surface-form match and a predicate-pattern
semantic match.
Our approach has a simple architecture, but
it outperforms the state-of-the-art, a system that has a
much more
complicated structure. 

\enote{hs}{given it's not clear you can do this, should we
  remove this?

We will release all the
code of this work, and the top-100 entity candidates of each
SimpleQuestions question to the community.

}

% Section \ref{sec:relatedwork} presents
% related work and \secref{taskdata} task and data, Section \ref{sec:linking} introduces our
% entity linker, Section \ref{sec:selection} elaborates our
% CNN model to match a tuple candidate with the question,
% Section \ref{sec:experiment} gives experimental results, and
% finally we conclude. 

\section{Related Work}\label{sec:relatedwork}
As mentioned in Section \ref{sec:intro}, factoid QA against
Freebase can be categorized into single-relation QA and
multi-relation QA. Much work has been done on multi-relation QA in
the past decade, especially after the release of benchmark
WebQuestions \cite{berant2013semantic}. Most state-of-the-art approaches
\cite{berant2013semantic,yahya2013robust,yao2014information,yih2015semantic}
are based on semantic parsing, where a question is mapped to
its formal meaning representation (e.g., logical form) and
then translated to a knowledge base (KB) query. The answers to the
question can then be retrieved simply by executing the
query. Other approaches retrieve a set
of candidate answers from KB using relation extraction
\cite{yao2014information,yih2014semantic,yao2015lean,bast2015more}
or distributed representations
\cite{bordes2014question,dong2015question,xu2016enhancing}. Our
method in this work explores CNN to learn distributed
representations for Freebase facts and questions.

SimpleQA was first investigated in
\cite{fader2013paraphrase} through PARALEX dataset against
knowledge base Reverb
\cite{fader2011identifying}. \newcite{yih2014semantic}
also investigate PARALEX  by a  system with some similarity
to ours
 -- they employ CNNs to match entity-mention and
predicate-pattern. Our model differs in
two-fold. (i) They use the same CNN architecture based on a
word-hashing technique \cite{huang2013learning} for both
entity-mention and predicate-pattern matches. Each word is
first preprocessed into a count vector of
character-\emph{trigram} vocabulary, then forwarded into the
CNN as input. We treat entities and mentions as 
character sequences. Our char-CNN for entity-mention match
is more end-to-end without data preprocessing. (ii) We
introduce attentive maxpooling  for better
predicate-pattern match.

The latest benchmark SimpleQuestions in SimpleQA is
introduced by \newcite{bordes2015large}. 
\newcite{bordes2015large}
tackle this problem by an embedding-based QA system
developed under the framework of Memory Networks
\cite{WestonCB14,sukhbaatar2015weakly}. The setting of the
SimpleQA corresponds to the elementary operation of
performing a single lookup in the memory. They investigate
the performance of training on the combination of
SimpleQuestions, WebQuestions and Reverb training
sets. \newcite{golub2016character} propose a character-level
attention-based encoder-decoder framework to encode the
question and subsequently decode into (subject, predicate)
tuple.   Our model in this work is much simpler than
these prior systems. \newcite{dai2016cfo} combine a unified conditional probabilistic framework with deep recurrent neural
networks and neural embeddings to get state-of-the-art performance.

Treating SimpleQA as fact selection is inspired
by work on  answer selection (e.g.,
\newcite{yu2014deep}, \newcite{yin2015abcnn},
\newcite{santos2016attentive}) that
looks for the correct answer(s) from some
candidates for a given question.  The answer candidates in
those tasks are raw text, not structured information as
facts in Freebase are. We are also inspired by work that generates natural language questions
given knowledge graph facts \cite{seyler2015generating,serban2016generating}. It hints that there exists a kind of match between natural language questions and FB facts.

%To make the whole picture clearer, w

\section{Task Definition and Data Introduction}
\seclabel{taskdata}
We first describe the  Freebase \cite{bollacker2008freebase} and SimpleQuestions task \cite{berant2013semantic}.

Freebase is a structured knowledge base in which entities
are connected by predefined predicates or ``relations''. All predicates
 are directional, connecting from
the subject  to the object. A triple
(subject, predicate, object), denoted as ($h$, $p$, $t$),
describes a fact; e.g., (U.S. Route 2,
major\_cities, Kalispell) refers to  the fact that U.S. Route 2
runs through the city of Kalispell.

SimpleQuestions benchmark, a typical SimpleQA task,  provides a set of
single-relation questions; each question is accompanied by a ground truth
fact. The object entity in the fact is the answer by
default. The  dataset is split into train (75,910), dev
(10,845) and test (21,687)
sets. This benchmark
 also provides two subsets of Freebase: FB2M (2,150,604 entities, 6,701 predicates, 14,180,937 atomic facts), FB5M (4,904,397 entities, 7,523 predicates, 22,441,880 atomic facts). While
single-relation questions are easier to handle than
questions with more complex and multiple relations,
single-relation question answering is still far from being
solved. Even in this restricted domain there are a large
number of paraphrases of the same question. Thus,
the problem of mapping from a question to a particular
predicate and entity in Freebase is hard.

% \begin{table}[t]
% \setlength{\belowcaptionskip}{-15pt}
% \setlength{\abovecaptionskip}{-5pt}
% \begin{center}
% \setlength{\tabcolsep}{2mm}
% \begin{tabular}{l|rr}
%  & FB2M  & FB5M \\\hline
% entities &2,150,604 & 4,904,397\\ 
% predicates & 6,701 & 7,523\\
% atomic facts & 14,180,937 & 22,441,880
% %facts (grouped) & 10,843,106 & 12,010,500
% \end{tabular}
% \end{center}
% \caption{\textbf{Knowledge Bases} used in this paper. FB2M and FB5M are two versions of Freebase.}\label{tab:fb} 
% \end{table}

The task assumes that single-relation questions can be
answered by querying a knowledge base such as Freebase with
a single subject and predicate argument. Hence, only the
tuple ($h$, $p$) is used to match the question. The
evaluation metric is accuracy.
Only
a fact that matches the ground truth in both
subject and predicate is counted as correct.

% WebQuestion contains simple questions as well as
% multi-predicate questions. But 86\% questions of it can be
% answered by a single fact \cite{bordes2015large}. The main
% contribution of this work is to propose a simple and
% strong system for single-predicate QA, but we also apply
% our system into the WebQuestion to see how far we can go
% in this more challenging task.

\section{Entity Linking}\label{sec:linking}
Given a question, the entity linker provides a set of  top-$N$ entity
candidates. 
The input of our deep learning model are (subject,
predicate) and (mention, pattern) pairs. 
Thus, given a question, two problems we have to solve are (i)
identifying candidate entities in Freebase that the question
refers to and (ii) identifying the span (i.e., mention) in the question that
refers to the entity. Each problem can be handled before the other, which results in  two entity linkers. (i) \textbf{Passive Entity Linker}: First search for entity candidates \emph{by all question words}, then use returned entities to guide the mention detection; (ii) \textbf{Active Entity Linker}: First identify the entity mention in the question, then \emph{use the mention span} to search for entity candidates. We now introduce them in detail.

\paragraph{Passive Entity Linker.} We perform 
entity linking
 by deriving 
the \emph{longest
consecutive common subsequence} (LCCS) between a question and entity
candidates and refer to it as $\sigma$. Given a question $q$ and all entity
names from Freebase, we perform the following three steps.

(i) Lowercase/tokenize  entity names and question

(ii) Use each component word of $q$ to retrieve
entities whose names contain this word. 
We refer to the set of all these entities as $C_e$.

%Compute three scores: $a=\frac{|\sigma|}{|q|}$,
%$b=\frac{|\sigma|}{|e|}$ and $c=\frac{p}{|q|}$. Finally, score 

(iii) For each entity candidate 
$e$ in $C_e$,
compute its LCCS $\sigma$ with the question $q$. Let
$p$ be the position of the last token of $\sigma$ in $q$.
Compute  $a=|\sigma|/|q|$,
$b=|\sigma|/|e|$ and $c=p/|q|$ where $|\cdot|$ is length in words. Finally, entity candidate $e$ is scored by the weighted sum $s_e=\alpha
a+\beta b+(1-\alpha-\beta)c$. Parameters $\alpha$ and $\beta$ are
tuned on dev. Top-$N$ ranked entities are kept for
each question.

\paragraph{Discussion.} Factor $a=|\sigma|/|q|$ means we prefer
 candidates that cover \emph{more consecutive words} of
the question. Factor $b=|\sigma|/|e|$ means that we prefer the candidates that cover \emph{more consecutive words
of the entity}. Factor $c=p/|q|$ means that
we prefer candidates that appear \emph{close to the end
of the question}; this is based on the
observation that most entity mentions are
far from the  beginning of the question. Despite the simplicity of this passive
entity linker, it outperforms other state-of-the-art entity
linkers of this SimpleQuestions task by a big
margin. Besides, this entity linker is  unsupervised
and runs fast. We will show its promise and investigate the
individual contributions of the three factors in
experiments.

Each question $q$
is provided top-$N$ entity candidates from Freebase by
entity linker. Then for \emph{mention detection}, we first compute the LCCS $\sigma$ \emph{on word level} between $q$
and entity $e$. If  the entity is longer than $\sigma$ and
has $l$ (resp.\ $r$) words on the left (resp.\ right) of
$\sigma$, then we extend $\sigma$ in the
question by $l$ left (resp.\ $r$ right) words  and select
this subsequence  as the candidate
mention. For example, entity ``U.S. Route 2'' and question
``what major cities does us route 2 run through'' have
LCCS $\sigma$
``route 2''.  The FB entity ``U.S. Route 2'' has one extra word
``u.s.'' on the left of $\sigma$, so we extend $\sigma$ by one left
word and the candidate mention is
``us route 2''. We do this so
that the mention has the same word size as the entity
string.\footnote{Only using LCCS as mention performed worse.}

In rare cases that the LCCS on the word level has length 0, we
treat both entity string and question as character sequence,
then compute LCCS $\sigma$ on character level. Finally, mention is
formed by expanding $\sigma$ on both sides up to a space or the
text boundary.

For each question, this approach to mention detection usually produces \emph{more
  than one
(mention, pattern) pair}.

\paragraph{Active Entity Linker.} 
In the training set of SimpleQuestions, the topic entity of each question is labeled. Active entity linker is then achieved by detecting mention in a question by sequential labeling. The key idea is to train a model to predict the text span of the topic entity which can match  the gold entity. This is inspired by some prior work. For example, \newcite{dai2016cfo} map the gold entity back to the text to label the text span for each question and then train a BiGRU-CRF model to do the mention detection. \newcite{golub2016character} propose a generative model which generates the topic entity based on character-level text spans with soft attention scores. 
Similar to the work \cite{dai2016cfo}, we trained a BiLSTM-CRF model to detect the entity mentions.

This approach to mention detection  produces \emph{only one
(mention, pattern) pair} for each question. Then, based on this detected mention, \emph{we use each word of it} to search for the entity candidates via the three steps in ``Passive Entity Linker''.

We presented two styles of mention detection in questions -- passive or active. In passive mention detection, the mention of a question depends on the entity candidates returned by an entity linker. Due to the different furface-forms of entity candidates, a question can be detected in different spans as mentions. Instead, active mention detection is conducted in a similar way with Name Entity Recognition. Hence, the mention does not depend on the returned entity candidates, a single-relation question has only one mention. Our experiments will show that active entity linker bring better coverage of ground truth entities, nevertheless this method requires the availability of entity-labeled questions as training data.

After mention detection, we then convert the question into the
tuple (mention, pattern) where pattern is created by
replacing the mention in the question with 
$<$e$>$.

\begin{figure}[t] 
\centering 
\subfigure[The whole system. Question: what major cities does us route 2 run through; Tuple: (``u.s. route 2'', ``major\_cities'')] { \label{fig:whole} 
\includegraphics[width=12cm]{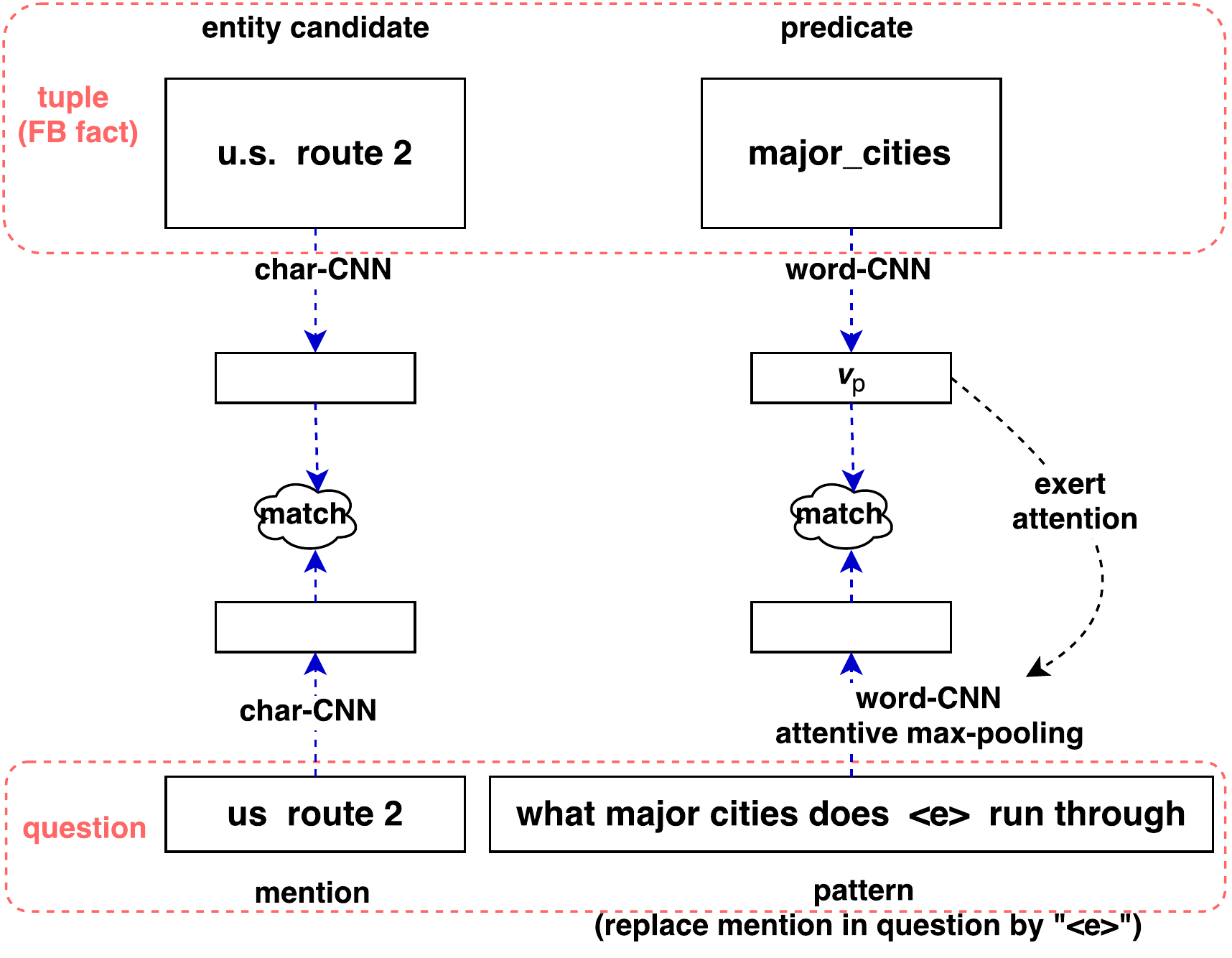} 
} 
\hspace{.1in}
\subfigure[Convolution for representation learning] { \label{fig:cnn} 
\includegraphics[width=3cm]{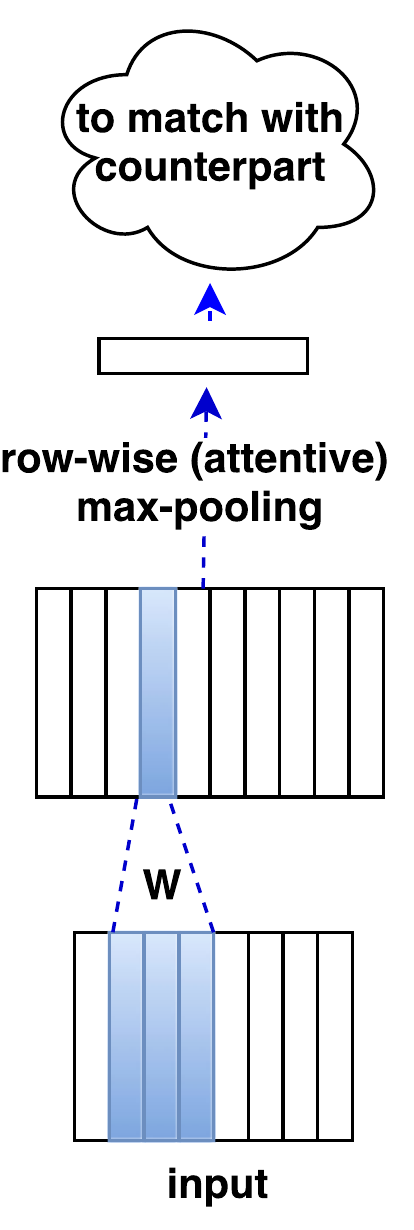} 
} 
\caption{CNN System for SimpleQA} 
\label{fig:wholeandcnn} 
\end{figure}
\section{Fact Selection}\label{sec:selection}
Entity linker provides top-$N$ entity candidates for each
question. All facts having those entities as subject form a
fact pool, then we build the system to seek the best. 

Our whole system is depicted in Figure \ref{fig:whole}. It consists of match from two aspects: (i) a CNN on character level (char-CNN) to detect the similarity of entity string and the mention string in surface-form (the left column); (ii) a CNN with attentive maxpooling (AMP) in word level (word-AMPCNN) to detect if the predicate is a paraphrase of the pattern. 

% cut forspace
%Specifically, the CNN with attentive maxpooling is designed
%to detect the match of predicate and pattern more
%: predicate string usually
%contains only a couple of words (one or two words in most
%cases) such as ``major cities'', and the pattern, such as
%``what major cities does <e> run through'', is a natural
%sentence containing some keywords ``major cities'' and
%less-informative words``what'' and ``run
%through''.5coherently. Our

Word-AMPCNN is motivated by the
observation  that
the FB predicate name is short and fixed whereas the corresponding
pattern in the question is highly variable in length and
word choice. Our hypothesis is that the predicate-pattern
match is best done based on keywords in the pattern (and
perhaps humans also do something similar) and that the CNN
therefore should identify helpful keywords.
Traditional maxpooling treats \emph{all
n-grams  equally}. In this work, we
propose \emph{attentive maxpooling} (AMP). AMP gives
\emph{higher weights to
n-grams that better match the predicate}. As a result, the
predicate-pattern match computed by the CNN is more likely
to be correct.

Next, we introduce the  CNN combined with
maxpooling for both char-CNN and word-CNN, then present 
AMPCNN. 
\figref{cnn} shows the common framework of char-CNN and
word-CNN; only 
input granularity and
maxpooling are different.

\subsection{Framework of CNN-Maxpooling}
Both char-CNN and word-CNN have two weight-sharing CNNs, as
they model two pieces of text. In what follows, we use
``entry'' as a general term for both character and word.

The \textbf{input layer} is  a sequence of entries of length
$s$ where each
entry is represented by
a $d$-dimensional randomly initialized embedding; thus
the sequence is represented as a feature map of
dimensionality  $d \times s$. 
Figure \ref{fig:cnn} shows the input layer as
the lower rectangle with multiple columns.

\textbf{Convolution Layer}  is used for  representation
learning from sliding $n$-grams. For an input sequence with $s$ entries: $v_1,v_2,\ldots,v_s$,
let vector $\mathbf{c}_i\in\mathbb{R}^{nd}$ be the
concatenated embeddings of $n$ entries
$v_{i-n+1},\ldots,v_{i}$ where $n$ is the filter width and $0< i
<s+n$.  Embeddings for  $v_i$, $i<1$ or $i>s$, are
zero padded.  We then generate the representation
$\mathbf{p}_i\in\mathbb{R}^d$ for the $n$-gram
$v_{i-n+1},\ldots,v_{i}$ using the convolution weights
$\mathbf{W}\in\mathbb{R}^{d\times nd}$:
\begin{equation}\label{eq:cnn}
\mathbf{p}_i=\mathrm{tanh}(\mathbf{W}\cdot\mathbf{c}_i+\mathbf{b})
\end{equation}
where bias $\mathbf{b}\in\mathbb{R}^d$. 

\textbf{Maxpooling.}
All $n$-gram representations $\mathbf{p}_i$ $(i=1\cdots s+n-1)$ are used to generate the  representation of input sequence $\mathbf{s}$ by maxpooling: $\mathbf{s}_j=\mathrm{max}(\mathbf{p}_{j1}, \mathbf{p}_{j2}, \cdots)$ ($j=1,\cdots, d$).

\begin{figure}[t]
\centering
\includegraphics[width=10cm]{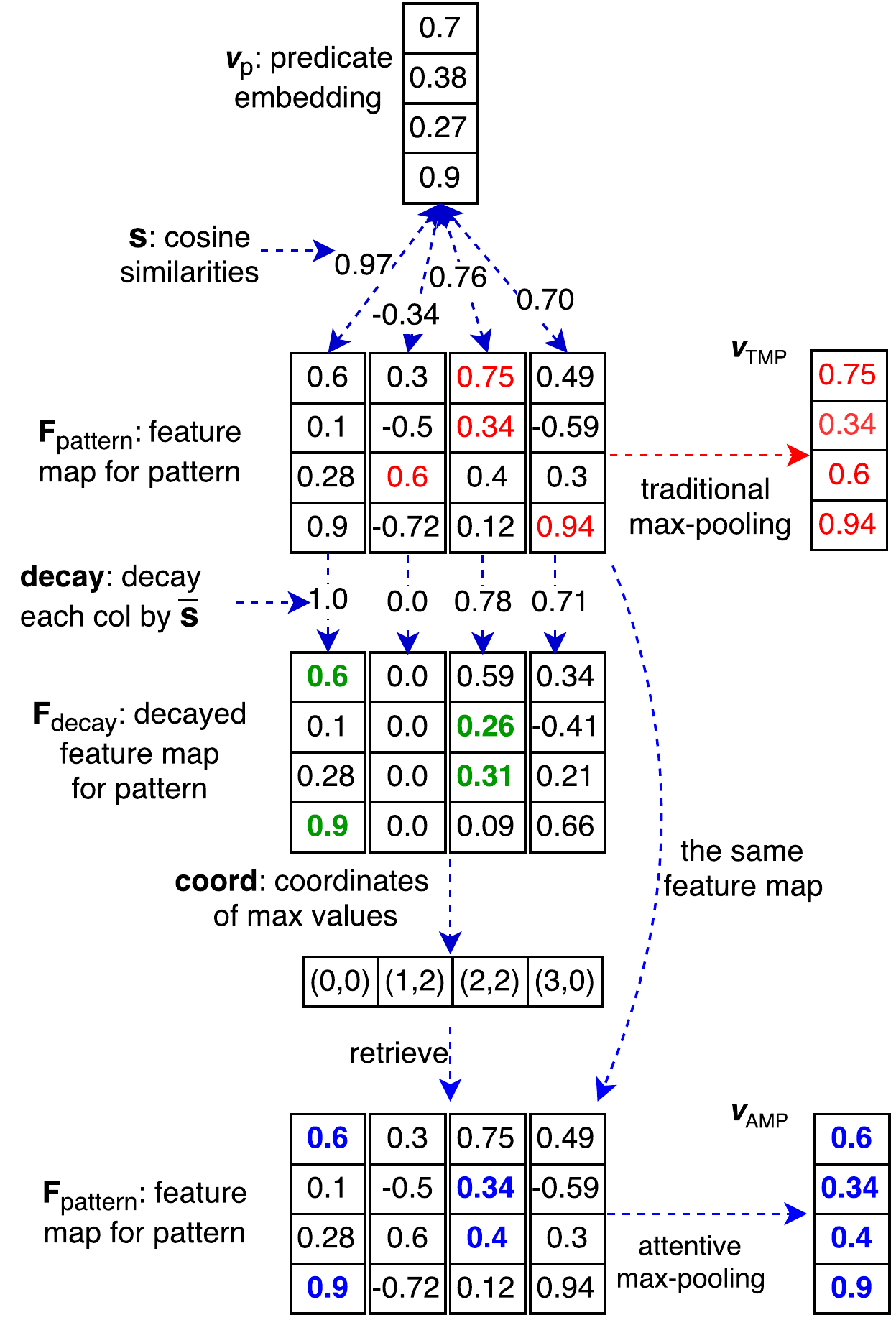}
\caption{Traditional maxpooling vs. Attentive maxpooling} \label{fig:pool}
\end{figure}

\subsection{AMPCNN: CNN-Attentive-Maxpooling}
\figref{pool} shows TMP (Traditional MaxPooling) and AMP
(Attentive MaxPooling) as we apply them to SimpleQA. Recall
that we use standard CNNs to produce (i)
the predicate representation $\mathbf{v}_p$ (see
\figref{whole})
and (ii) 
a feature
map of the pattern, i.e., a matrix with columns denoting $n$-gram
representations (shown in \figref{cnn}, the matrix below
``row-wise (attentive) maxpooling'').
In \figref{pool}, we
refer to the feature map as $\mathbf{F}\dnrm{pattern}$ and
to the predicate representation as $\mathbf{v}_p$.

TMPCNN, i.e., traditional maxpooling, outputs the vector
shown as $\mathbf{v}\dnrm{TMP}$; 
the same  $\mathbf{v}\dnrm{TMP}$ is produced for different 
$\mathbf{v}_p$.
The basic idea of AMPCNN is
to let the predicate $\mathbf{v}_p$ \emph{bias the selection and
weighting of subsequences of
the question to compute the representation of the
pattern}. The first step in doing that is to
compute similarity scores $\mathbf{s}$ 
between
the predicate representation $\mathbf{v}_p$ and
each column
vector
of $\mathbf{F}\dnrm{pattern}$:
\begin{equation}
\mathbf{s}_i=\mathbf{cos}(\mathbf{v}_p, \mathbf{F}\dnrm{pattern}[:,i])
\end{equation}
These cosines are then transformed into \emph{decay} values
by setting negative values to 0 (negatively correlated
column vectors are likely to be unrelated to the predicate)
and normalizing the positive values by dividing them by the
largest cosine (.97 in this case), so that the largest decay
value is 1.0. This is shown as  ``decay'' 
and $\overline{\mathbf{s}}$ in the figure.
Finally, we compute the reweighted feature map
$\mathbf{F}\dnrm{decay}$ as follows:
\begin{equation}
\mathbf{F}\dnrm{decay}[:,i]=\mathbf{F}\dnrm{pattern}[:,i]*\overline{\mathbf{s}}_i
\end{equation}

\eat{
Whereas features in $\mathbf{F}\dnrm{pattern}$ are
context-independent,
features in $\mathbf{F}\dnrm{decay}$ depend on the predicate:
features that are ``relevant'' to the predicate are kept and
highly weighted; nonrelevant features are discarded.

Finally,
 $\mathbf{v}\dnrm{AMP}$ is produced by standard maxpooling
from 
$\mathbf{F}\dnrm{decay}$ (the same way that
 $\mathbf{v}\dnrm{TMP}$ is produced
from 
$\mathbf{F}\dnrm{pattern}$).
}

In $\mathbf{F}\dnrm{decay}$, the matrix with four
green values, we
 can locate the maximal values in each
dimension. Notice that they are
not the true features by CNN any more, instead, \emph{they  convey the original
feature values as well as their importance to be
considered}. In $\mathbf{F}\dnrm{decay}$, we can see that the maximal values in each
dimension mostly come from the first column and the third
column which have relatively higher similarity scores 0.97
and 0.76 respectively to the predicate. \emph{We use the
coordinates of those maximal values  to retrieve features from  $\mathbf{F}\dnrm{pattern}$}
as a final pattern representation $\mathbf{v}\dnrm{AMP}$,  the blue column vector\footnote{We tried max-pooling over $\mathbf{F}\dnrm{decay}$ as $\mathbf{v}\dnrm{AMP}$ directly, but much worse performance was observed. }.

In summary, TMP has no notion of context. The novelty of AMP
is that it is guided by attentions from the context, in this case
attentions from the predicate. In contrast to TMP, we expect
AMP to mainly extract features that come from n-grams that
are related to the predicate.

\section{Experiments}\label{sec:experiment}

\subsection{Training Setup}
Our fact pool consists of all facts whose subject entity 
is in the top-$N$  entity candidates.
For train, we  sample 99 negative
facts for each ground truth fact; for dev and test, all fact candidates
are kept.

Figure \ref{fig:whole} shows two-way match between a tuple
 $t$ and a question $q$: entity-mention match by
char-CNN (score $m_e$), predicate-pattern match by word-AMPCNN
(score $m_r$). The overall ranking score of the pair is
$s_t(q,t)=m_e+m_r+s_e$
where $s_e$ is
the entity ranking score in entity linking phase.

Our objective is to minimize ranking loss:
\begin{equation}
\eqlabel{rankingloss}
l(q,t^+, t^-)=\mathrm{max}(0, \lambda+s_t(q,t^-)-s_t(q,t^+))
\end{equation}
where $\lambda$ is a constant.

We build word  and character vocabularies on
train.  OOV words and characters from dev and test
are mapped to an OOV index. Then,
words (resp.\ characters) are randomly initialized into $d\dnrm{word}$-dimensional
(resp.\ $d\dnrm{char}$-dimensional) embeddings.
The output dimensionality in convolution, i.e.,
Equation \ref{eq:cnn}, is the same as input
dimensionality. We employ Adagrad \cite{duchi2011adaptive},
$L_2$ regularization and diversity regularization
\cite{xie2015generalization}. Hyperparameters  (Table \ref{tab:hyper}) are tuned
on dev.
For active mention detection, we trained a two-layer BiLSTM followed by a CRF, the hidden layer sizes of both BiLSTM are 200.

\begin{table}[t]
\begin{center}
\begin{tabular}{ccccccc}
 $d\dnrm{word}$& $d\dnrm{char}$  & lr  & $L_2$ & $div$ & $k$ & $\lambda$\\\hline
500&100 & 0.1  & .0003 & .03 & [3,3]&0.5
\end{tabular}
\end{center}
\caption{Hyperparameters. 
$d\dnrm{word}$/$d\dnrm{char}$: embedding
dimensionality; lr: learning rate;   $L_2$: $L_2$ normalization; $div$: diversity regularizer; $k$: filter width in char/word-CNN. $\lambda$: see \eqref{rankingloss}}\label{tab:hyper} 
\end{table}

\subsection{Entity Linking}
In Table \ref{tab:entitylinking}, we compare our (passive and active) entity linkers with the state-of-the-art entity linker \cite{golub2016character} in this SimpleQA task. \newcite{golub2016character} report the coverage of ground truth by top-$N$ cases ($N\in\{1, 10, 20, 50, 100\}$). In addition, they explore a reranking algorithm to refine the entity ranking list. 

% \emph{We do not have reranking step in this work}.

Table \ref{tab:entitylinking} first shows  the \emph{overall}
performance of our passive entity linker and its performance without
factor $a$, $b$ or $c$ (\mbox{-a}, -b, -c).  Our passive entity linker outperforms
the baseline's \emph{raw} results by big margins and is
2--3 percent above their \emph{reranked} scores. This shows the
outstanding performance of our passive entity linker despite its
simplicity. The table also shows
that all three
factors ($a$, $b$, $c$) matter. Observations:
(i) Each
factor matters more when  $N$  is smaller.  This
makes sense because when $N$ reaches the entity vocabulary
size, all methods will have coverage 100\%. (ii) 
The position-related factor $c$ has less influence. From
top1 to top100, its contribution decreases from 4.3 
to .9. Our  linker still outperforms the
reranked baseline for $N\geq 20$. (iii) Factor $a$ is
dominant for small $N$, presumably because it chooses the
longer one when two candidates exist, which is critical for small $N$.
(iv) Factor $b$ plays a more
consistent role across different $N$.

The last column of Table \ref{tab:entitylinking} shows the overall results of our active entity linker, which are significantly better than the results  of baseline linker and our passive linker.
\emph{We release our entity linking results  for follow-up work to make better comparison.}

\subsection{SimpleQuestions}
Table \ref{tab:overallresult} compares AMPCNN
 with two baselines.
(i) \textbf{MemNN} \cite{bordes2015large}, an implementation
of memory network for SimpleQuestions task. (ii)
\textbf{Encoder-Decoder} \cite{golub2016character}, a
character-level, attention-based encoder-decoder LSTM
\cite{hochreiter1997long} model. (iii) \textbf{CFO} \cite{dai2016cfo}, the state-of-the-art system in this task with CNN or BiGRU subsystem. 

% Both MemNN and
% Encoder-Decoder
% use
% \emph{multiple data resources} (WebQuestion, SimpleQuestions
% and Paraphrases \cite{fader2014open}) and
% they are \emph{ensembles} with improvements over non-ensemble single
% systems
% from 62.2 to 63.9 on FB5M
% \cite{bordes2015large} 
% and from 65.9 to 66.2 on FB2M
% \cite{golub2016character}.
% Our system
% outperforms both baselines significantly even though it uses
% \emph{only a single dataset}
%   (SimpleQuestions) and uses \emph{no ensemble}.

\begin{table}[t]
% \setlength{\abovecaptionskip}{0pt} 
% \setlength{\belowcaptionskip}{-15pt} 
% \small
\begin{center}
\begin{tabular}{r|rr|crrr|c}
% &&\multicolumn{6}{c|}{MCTest-150}  & \multicolumn{6}{c}{MCTest-500} \\ 
&  \multicolumn{2}{c|}{\textbf{baseline}}  & \multicolumn{5}{c}{\textbf{Ours}}  \\ 
$\mathbf{N}$& \textbf{raw}&\textbf{rerank}& \textbf{passive-linker} &\textbf{-a}&\textbf{-b}&\textbf{-c} & \textbf{active-linker}\\\hline
1 & 40.9 & 52.9  & \textbf{56.6} & 11.0 & 34.9 & 52.3 & \textbf{73.6}\\
5 & -- & --  &  \textbf{71.1} & 29.5 & 49.5& 67.7 & \textbf{85.0} \\
10 & 64.3  & 74.0  & \textbf{75.2}  &40.7 &56.6 & 72.8 & \textbf{87.4}\\ 
20 & 69.3 & 77.8 & \textbf{81.0}  & 63.3& 62.4 & 78.6 & \textbf{88.8}\\
50 & 75.7  & 82.0  & \textbf{85.7}   &  77.1&67.1 &84.2 & \textbf{90.4}\\ 
100 & 79.6 & 85.4  & \textbf{87.9}  &81.2& 70.4 & 87.0 & \textbf{91.6}
\end{tabular}
\end{center}
\caption{Experimental results for entity linking}\label{tab:entitylinking} 
\end{table}

We report results for both passive entity linker and active entity linker. Furthermore, we  
compare AMPCNN to TMPCNN, i.e., 
we remove  attention and
representations for the
 predicate-pattern match are computed without
 attention. 
We choose
top-20 (i.e., $N=20$) entities returned by entity
linker.  Table \ref{tab:overallresult} shows that
AMPCNN with active  entity linker has optimal
performance for FB2M and
FB5M. Performance on
FB5M is slightly lower than   on FB2M, which should be
mainly due to the lower coverage for entity linking on FB5M
-- about 2\% below that on FB2M. In addition, our CNN
can still get competitive performance even if the
attention mechanism is removed (TMPCNN result). This hints that
CNN is  promising for SimpleQA.
\begin{table}[t]
% \setlength{\abovecaptionskip}{0pt} 
% \setlength{\belowcaptionskip}{-15pt} 
%  \small
\begin{center}
\begin{tabular}{lll|r@{\hspace{0cm}}rr@{\hspace{0cm}}r}
\textbf{Settings}&\multicolumn{2}{c|}{\textbf{Methods}}  & \textbf{FB2M}& &\textbf{FB5M}\\\hline
\multirow{6}{*}{{\footnotesize passive entity linker}} & \multirow{4}{*}{{\footnotesize baseline}}&random guess & 4.9 && 4.9&\\
&&MemNN    & 62.7 && 63.9&  \\
&&CFO w/ CNN    & - && 56.0 &  \\
&&CFO w/ BiGRU & - && 62.6&  \\\cline{2-6}
&\multirow{2}{*}{{\footnotesize CNN}} &TMPCNN   & \textbf{67.5}&$^*$  & \textbf{66.6}&$^*$  \\
&&AMPCNN    & \textbf{68.3}&$^*$  & \textbf{67.2}&$^*$  \\\hline \hline
%& CNN (no attention) & 67.5 && 66.6&\\\hline
\multirow{5}{*}{{\footnotesize active entity linker}} &
\multirow{3}{*}{{\footnotesize baseline}} & Encoder-Decoder  & 70.9 & & 70.3& \\
&& CFO w/ CNN & - && 71.1\\
&& CFO w/ BiGRU & - && 75.7 \\ \cline{2-6}
& \multirow{2}{*}{{\footnotesize CNN}}
% &AMPCNN FB-independent  & 63.9& & 62.1&\\
&TMPCNN   & 75.4&  & 74.6&\\ 
&&AMPCNN    & \textbf{76.4}&$^*$  & \textbf{75.9}&  
\end{tabular}
\end{center}
\caption{Experimental results for SimpleQuestions. Significant improvements over top baseline are marked with * (test of equal proportions, $p<.05$).}\label{tab:overallresult} 
\end{table}

% \begin{table}[t]
% \setlength{\abovecaptionskip}{0pt} 
% \setlength{\belowcaptionskip}{-15pt} 
%  \small
% \begin{center}
% % \setlength{\tabcolsep}{1mm}
% \begin{tabular}{ll|r@{\hspace{0cm}}rr@{\hspace{0cm}}r}
% \multicolumn{2}{c|}{methods}  & FB2M& &FB5M\\\hline
% \multirow{3}{*}{\rotatebox{90}{\footnotesize baseline}}&random guess & 4.9 && 4.9&\\
% &MemNN    & 62.7 && 63.9&  \\
% &Encoder-Decoder  & 70.9 & & 70.3& \\\hline
% %& CNN (no attention) & 67.5 && 66.6&\\\hline
% \multirow{2}{*}{\rotatebox{90}{\footnotesize CNN}}
% % &AMPCNN FB-independent  & 63.9& & 62.1&\\
% &TMPCNN FB-dependent  & 73.1&$^*$  & 71.6&$^*$\\ 
% &AMPCNN FB-dependent   & \textbf{74.0}&$^*$  & \textbf{72.3}&$^*$  
% \end{tabular}
% \end{center}
% \caption{Experimental results for SimpleQuestions. Significant improvements over ``Encoder-Decoder'' are marked with * (test of equal proportions, $p<.05$).}\label{tab:overallresult} 
% \end{table}

\enote{hs}{add back in if there is sapce
Recall that
we counted the value ``84.5\%'' by binary checking of
overlapping between ground truth entity string and detected
mention, this suggests that the true accuracy should be
lower than 84.5\%. Besides, even if we perfectly detected
84.5\% ground truth, considering further the coverage of
entity linker $\sim$80\%, the upper bound of the whole
system can not surpass 80\%$\times$84.5\% = 67.6\%.
}

% Interestingly, FB-independent  makes the
% performance worse by about 5\%.
% This can be attributed to the fact that FB-independent's
% accuracy of
% 84.5\% (\secref{mentiondetection})
% is not good enough for a  pipeline system.  
% FB-dependent can make the probability
% of mention detection for ground truth entities close to
% 100\% recall at the cost of 
% extracting false positive mentions (for subjects  of
% negative facts). But the mismatch of predicates in negative
% facts and patterns is expected to diminish the bad effect
% dramatically.

% Despite  the worse 
% (albeit still better than MemNN on FB2M)
% performance of FB-independent,
% it is more
% intuitive than  FB-dependent  because each question
% should have one mention in SimpleQuestions and this mention
% should not vary in dependence of  entity
% candidates that were falsely linked. What we need in the future is a more advanced
% approach for mention detection or NER.\footnote{Stanford NER
%   \cite{finkel2005incorporating} performed worse.}

% We hope this result provides lesson and clue
%for future work.

\subsection{Effect of Attentive Maxpooling (AMP)}
We compare AMP (one main contribution of this work) with three
CNN attention mechanisms that are representative of related work
in modeling two pieces of text: (i) \textbf{HABCNN}: Hierarchical
attention-based CNN \cite{yin2016attention}; (ii)
\textbf{ABCNN}: Attention-based CNN \cite{yin2015abcnn};
(iii) \textbf{APCNN}: CNN with attentive pooling
\cite{santos2016attentive}. 

%These three attention mechanisms
%are representative for convolution-pooling network when
%modeling two pieces of text.

Since
attentive matching of predicate-pattern is only one
part of our jointly trained system, it is hard to judge whether or not an attentive CNN
performs better than alternatives.
We therefore create a relation
classification (RC) subtask to compare AMP with baseline
schemes directly. RC task is created based on
SimpleQuestions:
label each question (converted into a pattern first)
with the ground truth predicate;
all other predicates of the gold subject entity are labeled
as negative. The resulting datasets have sizes
72,239 (train), 10,310 (dev) and 20,610 (test). It is worth mentioning that this relation classification task is not unspecific to the SimpleQA task, as RC is actually the predict-pattern match part. Hence, this RC subtask can be viewed to check how well the predict-pattern subsystem performs within the whole architecture, and the effectiveness of various attention mechanisms is more clear.

In  the three baselines, two pieces of text apply
attention to each other.
We adapt them into \emph{one-way attention} (OWA) as AMP does in this work: fix predicate
representation, and use it to guide the learning of pattern
representation. To be specific, ABCNN first gets predicate
representation by mean pooling, then uses this representation
to derive similarity scores of each n-gram in pattern as
attention scores, finally averages all n-gram embeddings
weighted by attentions as pattern representation. HABCNN
first gets predicate representation by max pooling, then
computes attention scores the same way as ABCNN, finally does
maxpooling over representations of top-$k$ similar n-grams.
APCNN is  similar to ABCNN except that the
similarity scores are computed by a nonlinear bilinear form.

\enote{hs}{reinstate if there is space
Its bilinear form brings
more parameters than ABCNN, HABCNN and the attentive
maxpooling in this work.
}

\begin{table}[t]
\begin{center}
% \setlength{\tabcolsep}{2mm}
% \small
\begin{tabular}{l|rr}
 &RC&Para \\\hline
 OWA-HABCNN \cite{yin2016attention}&.847 & 0\\
 OWA-ABCNN \cite{yin2015abcnn}&.902 & 0\\
 OWA-APCNN \cite{santos2016attentive}&.905& $\mathbb{R}^{d\times d}$\\\hline
 AMPCNN& .913&0 
\end{tabular}
\end{center}
\caption{Comparing different attention schemes of CNN in terms of  RC, \emph{extra} parameters brought (Para). }\label{tab:attention} 
\end{table}

Table \ref{tab:attention} shows that AMPCNN performs well on
relation classification, outperforming
the best baseline APCNN by 0.8\%. 
AMPCNN also 
has fewer parameters and
runs faster than APCNN.

\section{Conclusion}
This work explored CNNs for
the SimpleQA task. We made two main contributions. (i)
A simple and effective entity linker that brings
higher coverage of ground truth entities. (ii) An
attentive maxpooling stacked above convolution layer that
models the relationship between predicate and
question pattern more effectively. Our model shows 
outstanding performance on both simpleQA and
relation classification.

\textbf{Acknowledgments.} Wenpeng Yin and Hinrich
Sch\"{u}tze were partially supported by DFG (grant SCHU 2246/8-2).

\bibliography{coling2016}

\begin{thebibliography}{}

\bibitem[\protect\citename{Bast and Haussmann}2015]{bast2015more}
Hannah Bast and Elmar Haussmann.
\newblock 2015.
\newblock More accurate question answering on freebase.
\newblock In {\em Proceedings of CIKM}, pages 1431--1440.

\bibitem[\protect\citename{Berant \bgroup et al.\egroup
  }2013]{berant2013semantic}
Jonathan Berant, Andrew Chou, Roy Frostig, and Percy Liang.
\newblock 2013.
\newblock Semantic parsing on freebase from question-answer pairs.
\newblock In {\em Proceedings of EMNLP}, pages 1533--1544.

\bibitem[\protect\citename{Bollacker \bgroup et al.\egroup
  }2008]{bollacker2008freebase}
Kurt Bollacker, Colin Evans, Praveen Paritosh, Tim Sturge, and Jamie Taylor.
\newblock 2008.
\newblock Freebase: a collaboratively created graph database for structuring
  human knowledge.
\newblock In {\em Proceedings of SIGMOD}, pages 1247--1250.

\bibitem[\protect\citename{Bordes \bgroup et al.\egroup
  }2014]{bordes2014question}
Antoine Bordes, Sumit Chopra, and Jason Weston.
\newblock 2014.
\newblock Question answering with subgraph embeddings.
\newblock In {\em Proceedings of EMNLP}, pages 615--620.

\bibitem[\protect\citename{Bordes \bgroup et al.\egroup }2015]{bordes2015large}
Antoine Bordes, Nicolas Usunier, Sumit Chopra, and Jason Weston.
\newblock 2015.
\newblock Large-scale simple question answering with memory networks.
\newblock {\em arXiv preprint arXiv:1506.02075}.

\bibitem[\protect\citename{Dai \bgroup et al.\egroup }2016]{dai2016cfo}
Zihang Dai, Lei Li, and Wei Xu.
\newblock 2016.
\newblock {CFO}: Conditional focused neural question answering with large-scale
  knowledge bases.
\newblock In {\em Proceedings of ACL}, pages 800--810.

\bibitem[\protect\citename{Dong \bgroup et al.\egroup }2015]{dong2015question}
Li~Dong, Furu Wei, Ming Zhou, and Ke~Xu.
\newblock 2015.
\newblock Question answering over freebase with multi-column convolutional
  neural networks.
\newblock In {\em Proceedings of ACL-IJCNLP}, volume~1, pages 260--269.

\bibitem[\protect\citename{Duchi \bgroup et al.\egroup
  }2011]{duchi2011adaptive}
John Duchi, Elad Hazan, and Yoram Singer.
\newblock 2011.
\newblock Adaptive subgradient methods for online learning and stochastic
  optimization.
\newblock {\em JMLR}, 12:2121--2159.

\bibitem[\protect\citename{Fader \bgroup et al.\egroup
  }2011]{fader2011identifying}
Anthony Fader, Stephen Soderland, and Oren Etzioni.
\newblock 2011.
\newblock Identifying relations for open information extraction.
\newblock In {\em Proceedings of EMNLP}, pages 1535--1545.

\bibitem[\protect\citename{Fader \bgroup et al.\egroup
  }2013]{fader2013paraphrase}
Anthony Fader, Luke~S Zettlemoyer, and Oren Etzioni.
\newblock 2013.
\newblock Paraphrase-driven learning for open question answering.
\newblock In {\em Proceedings of ACL}, pages 1608--1618.

\bibitem[\protect\citename{Golub and He}2016]{golub2016character}
David Golub and Xiaodong He.
\newblock 2016.
\newblock Character-level question answering with attention.
\newblock In {\em Proceedings of EMNLP}.

\bibitem[\protect\citename{Hochreiter and Schmidhuber}1997]{hochreiter1997long}
Sepp Hochreiter and J{\"u}rgen Schmidhuber.
\newblock 1997.
\newblock Long short-term memory.
\newblock {\em Neural computation}, 9(8):1735--1780.

\bibitem[\protect\citename{Huang \bgroup et al.\egroup
  }2013]{huang2013learning}
Po-Sen Huang, Xiaodong He, Jianfeng Gao, Li~Deng, Alex Acero, and Larry Heck.
\newblock 2013.
\newblock Learning deep structured semantic models for web search using
  clickthrough data.
\newblock In {\em Proceedings of CIKM}, pages 2333--2338.

\bibitem[\protect\citename{LeCun \bgroup et al.\egroup
  }1998]{lecun1998gradient}
Yann LeCun, L{\'e}on Bottou, Yoshua Bengio, and Patrick Haffner.
\newblock 1998.
\newblock Gradient-based learning applied to document recognition.
\newblock In {\em Proceedings of the IEEE}, pages 2278--2324.

\bibitem[\protect\citename{Santos \bgroup et al.\egroup
  }2016]{santos2016attentive}
Cicero~dos Santos, Ming Tan, Bing Xiang, and Bowen Zhou.
\newblock 2016.
\newblock Attentive pooling networks.
\newblock {\em arXiv preprint arXiv:1602.03609}.

\bibitem[\protect\citename{Serban \bgroup et al.\egroup
  }2016]{serban2016generating}
Iulian~Vlad Serban, Alberto Garc{\'\i}a-Dur{\'a}n, Caglar Gulcehre, Sungjin
  Ahn, Sarath Chandar, Aaron Courville, and Yoshua Bengio.
\newblock 2016.
\newblock Generating factoid questions with recurrent neural networks: The 30m
  factoid question-answer corpus.
\newblock In {\em Proceedings of ACL}, pages 588--598.

\bibitem[\protect\citename{Seyler \bgroup et al.\egroup
  }2015]{seyler2015generating}
Dominic Seyler, Mohamed Yahya, and Klaus Berberich.
\newblock 2015.
\newblock Generating quiz questions from knowledge graphs.
\newblock In {\em Proceedings of WWW}, pages 113--114.

\bibitem[\protect\citename{Sukhbaatar \bgroup et al.\egroup
  }2015]{sukhbaatar2015weakly}
Sainbayar Sukhbaatar, Jason Weston, Rob Fergus, et~al.
\newblock 2015.
\newblock End-to-end memory networks.
\newblock In {\em Proceedings of NIPS}, pages 2431--2439.

\bibitem[\protect\citename{Weston \bgroup et al.\egroup }2015]{WestonCB14}
Jason Weston, Sumit Chopra, and Antoine Bordes.
\newblock 2015.
\newblock Memory networks.
\newblock In {\em Proceedings of ICLR}.

\bibitem[\protect\citename{Xie \bgroup et al.\egroup
  }2015]{xie2015generalization}
Pengtao Xie, Yuntian Deng, and Eric Xing.
\newblock 2015.
\newblock On the generalization error bounds of neural networks under
  diversity-inducing mutual angular regularization.
\newblock {\em arXiv preprint arXiv:1511.07110}.

\bibitem[\protect\citename{Xu \bgroup et al.\egroup }2016]{xu2016enhancing}
Kun Xu, Yansong Feng, Siva Reddy, Songfang Huang, and Dongyan Zhao.
\newblock 2016.
\newblock Enhancing freebase question answering using textual evidence.
\newblock {\em arXiv preprint arXiv:1603.00957}.

\bibitem[\protect\citename{Yahya \bgroup et al.\egroup }2013]{yahya2013robust}
Mohamed Yahya, Klaus Berberich, Shady Elbassuoni, and Gerhard Weikum.
\newblock 2013.
\newblock Robust question answering over the web of linked data.
\newblock In {\em Proceedings of CIKM}, pages 1107--1116.

\bibitem[\protect\citename{Yao and Van~Durme}2014]{yao2014information}
Xuchen Yao and Benjamin Van~Durme.
\newblock 2014.
\newblock Information extraction over structured data: {Q}uestion answering
  with {F}reebase.
\newblock In {\em Proceedings of ACL}, pages 956--966.

\bibitem[\protect\citename{Yao \bgroup et al.\egroup }2014]{yao2014freebase}
Xuchen Yao, Jonathan Berant, and Benjamin Van~Durme.
\newblock 2014.
\newblock Freebase {QA}: Information extraction or semantic parsing?
\newblock In {\em Proceedings of ACL Workshop on Semantic Parsing}, pages
  82--86.

\bibitem[\protect\citename{Yao}2015]{yao2015lean}
Xuchen Yao.
\newblock 2015.
\newblock Lean question answering over freebase from scratch.
\newblock In {\em Proceedings of NAACL-HLT}, pages 66--70.

\bibitem[\protect\citename{Yih \bgroup et al.\egroup }2014]{yih2014semantic}
Wen-tau Yih, Xiaodong He, and Christopher Meek.
\newblock 2014.
\newblock Semantic parsing for single-relation question answering.
\newblock In {\em Proceedings of ACL}, pages 643--648.

\bibitem[\protect\citename{Yih \bgroup et al.\egroup }2015]{yih2015semantic}
Wen-tau Yih, Ming-Wei Chang, Xiaodong He, and Jianfeng Gao.
\newblock 2015.
\newblock Semantic parsing via staged query graph generation: Question
  answering with knowledge base.
\newblock In {\em Proceedings of ACL}, pages 1321--1331.

\bibitem[\protect\citename{Yin \bgroup et al.\egroup }2016a]{yin2016attention}
Wenpeng Yin, Sebastian Ebert, and Hinrich Sch{\"u}tze.
\newblock 2016a.
\newblock Attention-based convolutional neural network for machine
  comprehension.
\newblock In {\em Proceedings of NAACL Human-Computer QA Workshop}.

\bibitem[\protect\citename{Yin \bgroup et al.\egroup }2016b]{yin2015abcnn}
Wenpeng Yin, Hinrich Sch{\"u}tze, Bing Xiang, and Bowen Zhou.
\newblock 2016b.
\newblock {ABCNN}: Attention-based convolutional neural network for modeling
  sentence pairs.
\newblock {\em TACL}.

\bibitem[\protect\citename{Yu \bgroup et al.\egroup }2014]{yu2014deep}
Lei Yu, Karl~Moritz Hermann, Phil Blunsom, and Stephen Pulman.
\newblock 2014.
\newblock Deep learning for answer sentence selection.
\newblock {\em Proceedings of ICLR Workshop}.

\end{thebibliography}
\bibliographystyle{acl}

\end{document}